\documentclass[letterpaper]{article} 
\usepackage{aaai25}  
\usepackage{times}  
\usepackage{helvet}  
\usepackage{courier}  
\usepackage[hyphens]{url}  
\usepackage{graphicx} 
\usepackage{natbib}  
\usepackage{caption} 
\usepackage{algorithm}
\usepackage{algorithmic}
\usepackage{multirow}
\usepackage{amsmath}
\usepackage{booktabs}
\usepackage{newfloat}
\usepackage{listings}
\DeclareCaptionStyle{ruled}{labelfont=normalfont,labelsep=colon,strut=off} 
\floatstyle{ruled}
\newfloat{listing}{tb}{lst}{}
\floatname{listing}{Listing}
\title{Multi-modal and Multi-scale Spatial Environment Understanding for \\ Immersive Visual Text-to-Speech}
\author{
    Rui Liu\textsuperscript{\rm 1}\thanks{Corrposending Author.},
    Shuwei He\textsuperscript{\rm 1},
    Yifan Hu\textsuperscript{\rm 1},
    Haizhou Li\textsuperscript{\rm 2, 3}
}
\affiliations{

    \textsuperscript{\rm 1}Inner Mongolia University, China\\
    \textsuperscript{\rm 2}Shenzhen Research Institute of Big Data, School of Data Science, The Chinese University of Hong Kong, Shenzhen, China\\
    \textsuperscript{\rm 3}Department of Electrical and Computer Engineering, National University of Singapore, Singapore\\
    liurui\_imu@163.com, shuwei\_he@163.com, hyfwalker@163.com, haizhouli@cuhk.edu.cn
}

\usepackage{bibentry}

\begin{document}

\maketitle

\begin{abstract}

Visual Text-to-Speech (VTTS) aims to take the environmental image as the prompt to synthesize the reverberant speech for the spoken content.
The challenge of this task lies in understanding the spatial environment from the image.
Many attempts have been made to extract global spatial visual information from the RGB space of an spatial image. However, local and depth image information are crucial for understanding the spatial environment, which previous works have ignored.
To address the issues, we propose a novel multi-modal and multi-scale spatial environment understanding scheme to achieve immersive VTTS, termed M$^{2}$SE-VTTS.
The multi-modal aims to take both the RGB and Depth spaces of the spatial image to learn more comprehensive spatial information, and the multi-scale seeks to model the local and global spatial knowledge simultaneously.
Specifically, we first split the RGB and Depth images into patches and adopt the Gemini-generated environment captions to guide the local spatial understanding. 
After that, the multi-modal and multi-scale features are integrated by the local-aware global spatial understanding. 
In this way, M$^{2}$SE-VTTS effectively models the interactions between local and global spatial contexts in the multi-modal spatial environment.
Objective and subjective evaluations suggest that our model outperforms the advanced baselines in environmental speech generation.
The code and audio samples are available at: https://github.com/AI-S2-Lab/M2SE-VTTS.

\end{abstract}

%

\section{Introduction}

Visual Text-to-Speech (VTTS) aims to leverage the environmental image as the prompt to generate the reverberant speech that corresponds to the spoken content. 
With the advancement of human-computer interaction, VTTS has become integral to intelligent systems and plays an important role in fields such as augmented reality (AR) and virtual reality (VR) \cite{ViT-TTS}.

Unlike acoustic matching tasks that transform input speech to match the environmental conditions of a reference source \cite{vam, audioldm, svam, diffrent}, VTTS seeks to synthesize speech with the environmental characteristics of the reference based on given textual content \cite{ms2ku}.
For example, \citet{voiceldm} utilizes the pre-trained CLAP model to map a textual or audio description into an environmental feature vector that controls the reverberation aspects of the generated audio.
\citet{eatts} design an environment embedding extractor that learns environmental features from the reference speech.
In more recent studies, \citet{ViT-TTS} propose a visual-text encoder based on the transformer to learn global spatial visual information from the RGB image.
Building on these advancements, this paper focuses on employing visual information as the cue to generate reverberation audio for the targeted scene.

However, previous VTTS methods have not fully understood the spatial environment, due to the neglect of local and depth image information.
For example, the local elements in the spatial environment can directly influence the reverberation style. Specifically, the hard surfaces such as tables reflect sound waves, while softer materials like carpets absorb them, directly affecting the audio's authenticity and naturalness \cite{lavd, vam, ViT-TTS}.
In addition, the Depth space of the image contains positional relationships within the spatial environment \cite{avlea, lavd}, such as the arrangement of objects, the position of the speaker and the room geometry.
Therefore, it is crucial for VTTS systems to accurately capture the local and depth spatial environment information simultaneously.

To address the issues, we propose a novel \textbf{multi-modal} and \textbf{multi-scale} spatial environment understanding scheme to achieve immersive VTTS, termed \textbf{M$^{2}$SE-VTTS}.
The \textbf{multi-modal} aims to take both the RGB and Depth spaces of the spatial image to learn more comprehensive spatial information, such as the speaker's location and the positions of key objects that influence sound absorption and reflection.
The \textbf{multi-scale} seeks to model the impact of local and global spatial knowledge on reverberation.
Specifically, we first split the RGB and Depth images into patches following the visual transformer strategy \cite{vit}.
In addition, we adopt the Gemini-generated \cite{gemini} environment captions to guide the local spatial understanding based on an identification mechanism.
After that, the local-aware global spatial understanding takes the multi-modal and multi-scale features as input and progressively integrates spatial environment knowledge.
In this way, M$^{2}$SE-VTTS effectively models the interactions between local and global spatial contexts among the multi-modal spatial environment.
The main contributions of this paper include:
\begin{itemize}
\item We propose a novel multi-modal and multi-scale spatial environment understanding framework, termed M$^{2}$SE-VTTS, that leverages both the RGB and Depth information to enhance the synthesis of immersive reverberation speech.
\item Our approach comprehensively integrates both local and global spatial elements, providing a more comprehensive understanding of the spatial environment, which is crucial for accurately modeling environmental reverberation.
\item Objective and subjective experimental evaluations demonstrate that our model significantly outperforms all existing state-of-the-art benchmarks in generating environmental speech. 
\end{itemize}

\begin{figure*}[t!]
\centering
\centerline{
\includegraphics[width=1.02\linewidth]{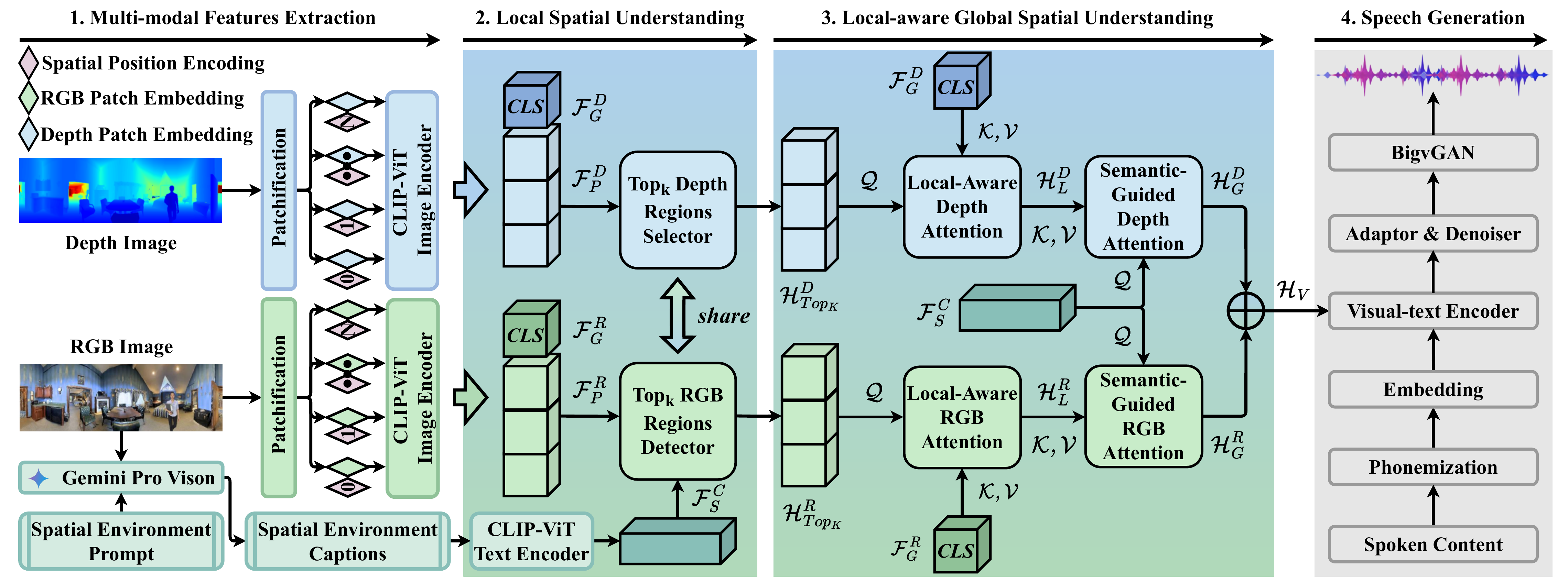}
}
\caption{The overall architecture of M$^{2}$SE-VTTS.}
\label{fig:model}
\end{figure*}


\section{Related Works}

\subsection{Spatial Environment Understanding}

Spatial environment understanding plays a crucial role in spatial cognition, particularly in complex three-dimensional scenes where accurate spatial comprehension is essential for applications such as robotic navigation, augmented reality, and autonomous driving.
In the visual domain, researchers often employ multi-modal and multi-scale approaches to capture and analyze spatial information more comprehensively \cite{soundspaces, segnext, seu1, symphonize, seu2, seu3}.
For instance, \citet{SpatialRGPT} enhances Vision Language Models (VLMs) by introducing a data curation pipeline and a plugin module that improves the understanding of 3D spatial relationships by integrating depth information and learning regional representations from 3D scene graphs.
\citet{Scene-llm} further extends the capabilities of language models by incorporating 3D visual data, improving the embodied agents' reasoning and decision-making abilities in interactive 3D environments.
This approach is particularly effective in tasks such as dense annotation and interactive planning.
Similarly, \citet{adaptive} addresses the challenge of depth estimation in autonomous driving by proposing a method that fuses single-view and multi-view depth estimates, showing robust performance in scenarios with sparse textures and dynamic objects.
These studies demonstrate that multi-modal and multi-scale methods are pivotal in advancing the comprehensive understanding of spatial environments, as they effectively integrate information from various sources to enhance spatial reasoning in complex environments.

While these works have made significant strides in improving spatial understanding in VLMs, they primarily focus on extracting global spatial information, often overlooking the importance of local and depth information.
Our work differs from these approaches in several key aspects: (1) We focus on the Visual VTTS task, rather than solely on visual spatial reasoning; (2) Unlike the previous works, we emphasize the integration of both local and depth image information, in addition to global spatial data from RGB images, to achieve a more holistic understanding of the spatial environment.
These differences allow our approach to better address the unique challenges of the VTTS task and excel in complex spatial environments.

\subsection{LLM-based Image Understanding}
In recent years, leveraging Large Language Models (LLMs) for image understanding has emerged as a significant research focus in the fields of computer vision and natural language processing. 
By integrating the powerful natural language processing capabilities of LLMs with visual information, researchers have developed multi-modal large language models (MLLMs) to tackle complex tasks such as visual question answering, image captioning, and image comprehension \cite{OMGLLaVA, beyond}. 
For example, \citet{visionbylanguage} pioneered the combination of pre-trained vision encoders with language models, utilizing a Perceiver Resampler module to extract features from images for generating textual descriptions, thus achieving cross-modal image-text alignment. 
Building on this, \citet{X-Former} introduced a Querying Transformer (Q-Former) that extracts the most relevant visual features through cross-modal fusion, enhancing the model's visual understanding capabilities. 
\citet{meerkat} further advanced these efforts by incorporating spatial coordinate information, thereby improving multi-modal models' abilities in object localization and visual reasoning. 
However, these approaches largely focus on capturing global visual features, which, while effective in many cases, show limitations when dealing with tasks that require fine-grained visual understanding. 
To address this challenge, \citet{X-Former} proposed a novel approach that combines contrastive learning (CL) with masked image modeling (MIM), integrating features from both CLIP-ViT \cite{clip} and MAE-ViT vision encoders. 
X-Former uses a dual cross-attention mechanism to align visual and language features, demonstrating superior performance in fine-grained visual tasks, such as object counting and fine-grained category recognition. 
In contrast, traditional multi-modal models often struggle with these tasks due to their limited ability to capture local details effectively.

While these methods have significantly advanced the capabilities of multi-modal visual understanding, they still face limitations when applied to specific spatial environment perception tasks. 
Our work introduces a novel multi-modal and multi-scale spatial environment understanding scheme, designed to overcome the shortcomings of existing models in capturing the local and depth information. 
Unlike previous approaches, our method integrates both RGB and depth information and utilizes Gemini-generated environmental captions to guide local spatial understanding. By fusing multi-modal and multi-scale features, our method provides a more comprehensive modeling of spatial environments, offering robust support for VTTS tasks, particularly in understanding the spatial layout and environmental characteristics of complex scenes.

\section{Methodology}

As shown in the pipeline of Fig. \ref{fig:model}, the proposed M$^2$SE-VTTS consists of four components: 1) Multi-modal Features Extraction; 2) Local Spatial Understanding; 3) Local-aware Global Spatial Understanding and 4) Speech Generation.
As mentioned previously, multi-modal features, including the RGB and Depth space representation of an image, can provide more comprehensive information about the spatial environment. 
To understand the interactions between local and global spatial contexts, the multi-modal and multi-scale knowledge is integrated by local-aware global spatial understanding. 
The following subsections provide detailed descriptions of the design and training processes for these components.

\subsection{Multi-modal Features Extraction}

Given the RGB and Depth image pairs of the spatial environment $\{\mathcal{V}_R, \mathcal{V}_D\}$, we first partition them into $M$ patches. 
In addition, we employ the image encoder of a pre-trained CLIP \cite{clip} model with frozen parameters to extract patch-level features as $\mathcal{F}_P^R$, $\mathcal{F}_P^D \in \mathcal{R}^{M \times D}$ from each of $\mathcal{V}_R$ and $\mathcal{V}_D$, where $D$ denotes the dimensionality of the features and $M$ indicates the number of patches per image. 
As illustrated in Fig \ref{fig:model}, a special $[CLS]$ token is used at the beginning of the first patch to represent the global-level features as $\mathcal{F}_G^R$, $\mathcal{F}_G^D \in \mathcal{R}^{1 \times D}$.

\subsection{Local Spatial Understanding}

As shown in the second panel of Fig. \ref{fig:model}, Local Spatial Understanding consists of three parts: 
1) LLM-based Spatial Semantic Understanding, leveraging Gemini's powerful multi-modal understanding capabilities to accurately convert complex visual scenes into semantic information; 
2) Top$_k$ RGB Regions Detector, guided by environmental captions to identify crucial semantic information of the RGB space of the image;
and 3) Top$_k$ Depth Regions Selector, selecting important semantic information of the Depth space of the image.

\subsubsection{LLM-based Spatial Semantic Understanding}

To capture rich spatial information, including the spatial positions of objects, their arrangement, and the overall scene structure, we utilize Gemini's advanced multi-modal understanding capabilities to convert the complex visual data into the structured caption. 
This approach enables us to accurately extract and represent the spatial semantics embedded within the image.

First of all, the spatial environment captions are generated using the Gemini Pro Vision, which is a multi-modal large language model configured with its default settings. 
The prompt designed for Gemini is as follows: ``Observe this panoramic image and briefly describe its content. Identify the objects in the image in one to two sentences, focusing only on key information and avoiding descriptive words." 
After analysis by Gemini, the spatial environment in Fig. \ref{fig:model} is described as follows: ``The image shows a spacious, circular room with a blue and white color scheme. It features a dining table with chairs, a kitchenette, a bedroom area with a bed, and a person standing in the center of the room." 
In the end, the caption $\mathcal{C}$ is tokenized into $N$ individual words, represented as $\mathcal{C}=\{c_ {n}\}_{n=1}^N$. 
And each word $c_n$ is represented as a fixed-length vector using word embeddings, which are input into the text encoder of a pre-trained CLIP model to obtain spatial semantic features $\mathcal{F}_S^C$, where $\mathcal{F}_S^C \in \mathcal{R}^{1 \times D}$. 
It is important to note that the $[CLS]$ token is used to aggregate and represent the overall semantic information of the entire input text, with this embedding vector serving as the primary representation of the text when aligning with image features.

\subsubsection{Top$_k$ RGB Regions Detector} 

Our goal is to identify and focus on the image regions that significantly influence sound propagation and reflection characteristics, enabling more accurate simulation of the reflection and absorption effects of different materials and surfaces, thereby making the generated speech more natural and realistic.

To begin with, we apply the spatial attention to $\mathcal{F}_{P}^{R}$ and $\mathcal{F}_{S}^{C}$ after using a linear projection layer, which is formalized as:
\begin{equation}
    \hat{\mathcal{F}}_{P}^{R}, \mathcal{A}_P^R = MultiHead(\mathcal{F}_{S}^{C}, \mathcal{F}_{P}^{R}, \mathcal{F}_{P}^{R}),
\end{equation}
where $\hat{\mathcal{F}}_{P}^{R}$ represents the updated features from $\mathcal{F}_{P}^{R}$, and $\mathcal{A}_P^R$ denotes the average attention weights across all heads, with $\mathcal{A}_P^R \in (0, 1)^{M}$.
Inspired by SRSM \cite{pstm}, after that, we introduce a detection operation, denoted as $\Phi_{LSU}$ to identify the patches with the highest Top$_k$ attention weights and their indices:
\begin{equation}
    \mathcal{H}_{Top_{k}}^{R}, \Omega_{R} = \Phi_{LSU}(\hat{\mathcal{F}}_{P}^{R}, \mathcal{A}_P^R, Top_{k}),
\end{equation}
where $\mathcal{H}_{Top_{k}}^{R} \in \mathcal{R}^{Top_{k} \times D}$ represents the detected local features of the RGB space, and $\Omega_{R} \in \{0, 1, \ldots, M \}^{Top_{k}}$ represents the indices corresponding to the highest Top$_{k}$ weights in $\mathcal{A}_P^R$.

\subsubsection{Top$_k$ Depth Regions Selector} 

This module aims to capture the relative distances of key objects, their arrangement, and the geometric layout of the room within the spatial environment, and to accurately simulate sound propagation and reflection, thereby generating reverberation that more closely aligns with the actual physical space. 
This module implements a selection attention-based strategy similar to $\Phi_{LSU}$.

Specifically, we take the indices $\Omega_{R}$ from $\Phi_{LSU}$ to select the corresponding crucial patch-level depth features. 
This approach is based on the following three key considerations:
1) the CLIP is pre-trained using RGB images paired with text, resulting in a stronger correlation between RGB and textual data compared to Depth information; 
2) by maintaining consistent patch indices across both RGB and Depth modalities, we ensure spatial coherence, allowing the model to accurately align and integrate features from the same spatial locations; 
and 3) this alignment further prevents potential information redundancy or conflicts between the modalities, ensuring that the model is better equipped to precisely capture and utilize complementary features from both RGB and Depth data. 
This process can be formulated as:
\begin{equation}
    \mathcal{H}_{Top_{k}}^{D} = \Psi_{LSU}(\mathcal{F}_{P}^{D}, \Omega_{R}),
\end{equation}
where $\mathcal{H}_{Top_{k}}^{D}$ represents the selected local features of the Depth space, and $\mathcal{H}_{Top_{k}}^{D} \in \mathcal{R}^{Top_{k} \times D}$.

\subsection{Local-aware Global Spatial Understanding}
As shown in the third panel of Fig. \ref{fig:model}, Local-aware Global Spatial Understanding aims to effectively model the interactions between local semantics and the global spatial context, which consists of two parts: 
1) the Local-aware RGB/Depth Attention, which focuses on learning the interactions between local details and global spatial features, 
and 2) the Semantic-Guided RGB/Depth Attention, which enhances the understanding of spatial contexts by integrating semantic information with the local-aware global features.

\subsubsection{Local-aware RGB/Depth Attention} 
This section aims to understand how local spatial details, such as the position and material of key objects, interact within the overall spatial layout and to comprehend the spatial relationships across different scales in the scene, thereby enabling the generation of reverberation that more accurately reflects the actual physical environment.

For \textbf{the RGB image}, given its $\mathcal{H}_{Top_{k}}^{R}$ and $\mathcal{F}_{G}^{R}$, we perform \textit{the Local-aware RGB Attention} to model the interactions between the local and global spatial knowledge of the RGB space after using a linear projection layer, which is formulated as follows:
\begin{equation}
    \mathcal{H}_{L}^{R} = MultiHead(\mathcal{H}_{Top_{k}}^{R}, \mathcal{F}_{G}^{R}, \mathcal{F}_{G}^{R}),
\end{equation}
where $\mathcal{H}_{L}^{R} \in \mathcal{R}^{Top_{k} \times D}$ is updated from $\mathcal{F}_{G}^{R}$. 

For \textbf{the Depth image}, giving its $\mathcal{H}_{Top_{k}}^{D}$ and $\mathcal{F}_{G}^{D}$, \textit{the Local-aware Depth Attention} adopts a similar strategy, which is formulated as follows:
\begin{equation}
    \mathcal{H}_{L}^{D} = MultiHead(\mathcal{H}_{Top_k}^{D}, \mathcal{F}_G^D, \mathcal{F}_G^D),
\end{equation}
where $\mathcal{H}_{L}^{D} \in \mathcal{R}^{Top_{k} \times D}$ is updated from $\mathcal{F}_{G}^{D}$.

\subsubsection{Semantic-Guided RGB/Depth Attention} 

To deepen our understanding of the complex relationships between spatial contexts across different scales and to enhance the model's performance in the multi-modal environment, we further employ a semantic-guided attention mechanism to achieve a more advanced fusion of local and global spatial features. 

For \textbf{the RGB image}, given its $\mathcal{H}_{L}^{R}$ and $\mathcal{F}_{S}^{C}$, we adopt \textit{the Semantic-Guided RGB Attention} to attain an advanced understanding between the local and global spatial contexts following a linear projection layer, which is formulated as follows:
\begin{equation}
    \mathcal{H}_{G}^{R} = MultiHead(\mathcal{F}_S^C, \mathcal{H}_{L}^{R}, \mathcal{H}_{L}^{R}),
\end{equation}
where $\mathcal{H}_{G}^{R} \in \mathcal{R}^{1 \times D}$ is updated from $\mathcal{F}_S^C$.

For \textbf{the Depth image}, \textit{the Semantic-Guided Depth Attention} employs a similar method to learn an advanced understanding of the Depth space, which is formulated as follows:
\begin{equation}
    \mathcal{H}_{G}^{D} = MultiHead(\mathcal{F}_S^C, \mathcal{H}_{L}^{D}, \mathcal{H}_{L}^{D})
\end{equation}

Eventually, we integrate the multi-modal and multi-scale features to derive a comprehensive representation of the spatial environment, which is formulated as follows:
\begin{equation}
    \mathcal{H}_V = \lambda_1 {\mathcal{H}}_{G}^{R} + \lambda_2 \mathcal{H}_{G}^{D},
\end{equation}
where the weights, $\lambda_1$ and $\lambda_2$, are both set to 0.5.

\subsection{Speech Generation}

As illustrated in Fig. \ref{fig:model}, we adopt ViT-TTS as the backbone for our TTS system.
To begin with, the phoneme embeddings and visual features are converted into hidden sequences.
In addition, the variance adaptor predicts the duration of each hidden sequence to regulate the length of the hidden sequences to match that of speech frames.
After that, different variances like pitch and speaker embedding are incorporated into hidden sequences following \citet{ren2021fastspeech}. 
Furthermore, the spectrogram denoiser iteratively refines the length-regulated hidden states into mel-spectrograms. 
In the end, the BigVGAN \cite{bigvgan} transforms mel-spectrograms into waveform. 
For more details, please refer to the ViT-TTS \cite{ViT-TTS}.

\section{Experiments and Results} 

\subsection{Dataset} 

We employ the SoundSpaces-Speech dataset \cite{lavd}, which is developed on the SoundSpaces platform using real-world 3D scans to simulate environmental audio.
To enhance the dataset, we refine it following the approach described in \citet{vam, ViT-TTS}.
Specifically, we exclude out-of-view samples and divide the remaining data into two subsets: test-unseen and test-seen. The test-unseen subset includes room acoustics derived from novel images, while the test-seen subset contains scenes previously observed during training.
The dataset consists of 28,853 training samples, 1,441 validation samples, and 1,489 testing samples.
Each sample includes clean text, reverberation audio, and panoramic camera RGB-D images.
To preprocess the text, we convert the sequences into phoneme sequences using an open-source grapheme-to-phoneme tool \footnote{https://github.com/Kyubyong/g2p}.

Following common practices \cite{fastspeech, prodiff, catts, ECSS}, we preprocess the speech data in three steps.
First, we extract spectrograms with an FFT size of 1024, a hop size of 256, and a window size of 1024 samples.
Next, we convert the spectrogram into a mel-spectrogram with 80 frequency bins.
Finally, we extract the F0 (fundamental frequency) from the raw waveform using Parselmouth \footnote{https://github.com/YannickJadoul/Parselmouth}.
These preprocessing steps ensure consistency with prior work and prepare the data for subsequent modeling.

\subsection{Implementation Details}

For the visual modality, we utilize the pre-trained CLIP-ViT-L/14 as the visual feature extractor. 
This model generates 768-dimensional feature vectors at both global and patch levels for each visual snippet. 
These visual features undergo a linear transformation and are subsequently aligned with the 512-dimensional hidden space of the phoneme embeddings. 
The phoneme vocabulary consists of 74 distinct phonemes.
The cross-modal fusion module employs two attention heads, while all other attention mechanisms use four heads each. The patch number, Top$_k$, is set to 140. The configuration of other encoder parameters follows the implementation in ViT-TTS.
In the denoiser module, we use five transformer layers with a hidden size of 384 and 12 heads. Each transformer block functions as the identity, with $T$ set to 100 and $\beta$ values increasing linearly from $\beta_1 = 10^{-4}$ to $\beta_T = 0.06$. 
This configuration facilitates effective noise reduction and enhances the quality of the generated outputs.

The training process consists of two stages. 
In the pre-training stage, we adopt the encoder pre-training strategy from ViT-TTS, training the encoder for 120k steps until convergence. 
In the main training stage, the M$^2$SE-VTTS model is trained on a single NVIDIA A800 GPU with a batch size of 48 sentences, extending over 160k steps until convergence. 
During inference, we use a pre-trained BigVGAN as the vocoder to transform the generated mel-spectrograms into waveforms. 
Further details on the model configuration and implementation are provided in Appendix A \footnote{https://shorturl.at/G0zpV}.

\begin{table*}[t]
\centering

\setlength{\baselineskip}{0.85mm}{
\begin{tabular}{l|ccc|ccc}
\hline
\multirow{2}{*}{\textbf{System}} & \multicolumn{3}{c|}{\textbf{Test-Unseen}}                                                                                                & \multicolumn{3}{c}{\textbf{Test-Seen}}                                                                                                   \\
                                 & \textbf{MOS} ($\uparrow$)                & \textbf{RTE} ($\downarrow$)    & \textbf{MCD} ($\downarrow$)    & \textbf{MOS}  ($\uparrow$)              & \textbf{RTE}  ($\downarrow$)   & \textbf{MCD} ($\downarrow$)  \\ \hline
GT                               & 4.353 $\pm $ 0.023                         & /                                            & /                                            & 4.348 $\pm $ 0.022                         & /                                            & /                                            \\
GT(voc.)                         & 4.149 $\pm $ 0.027                         & 0.0080                                       & 1.4600                                       & 4.149 $\pm $ 0.023                         & 0.0060                                       & 1.4600                                       \\ \hline
ProDiff \cite{prodiff}                          & 3.550 $\pm $ 0.023                          & 0.1341                                       & 4.7689                                       & 3.647 $\pm $ 0.023                         & 0.1243                                       & 4.6711                                       \\
DiffSpeech    \cite{diffsinger}                   & 3.649 $\pm $ 0.022                         & 0.1193                                       & 4.7923                                       & 3.675 $\pm $ 0.011                         & 0.1034                                       & 4.6630                                       \\
VoiceLDM       \cite{voiceldm}                  & 3.702 $\pm $ 0.020                         & 0.0825                                       & 4.8952                                       & 3.702 $\pm $ 0.025                         & 0.0714                                       & 4.6572                                       \\
ViT-TTS-ResNet18   \cite{ViT-TTS}              & 3.700 $\pm $ 0.025                         & 0.0759                                       & 4.5933                                       & 3.804 $\pm $ 0.022                         & 0.0677                                       & 4.5535                                       \\
ViT-TTS-CLIP     \cite{ViT-TTS}                & 3.651 $\pm $ 0.023                         & 0.0772                                       & 4.5871                                       & 3.746 $\pm $ 0.023                         & 0.0678                                       & 4.5385                                       \\ \hline
\textbf{M$^2$SE-VTTS}            & \textbf{3.849 $\pm $ 0.025}                & \textbf{0.0744}                              & \textbf{4.4215}                              & \textbf{3.939 $\pm $ 0.022}                & \textbf{0.0642}                              & \textbf{4.3809}                              \\ \hline
\end{tabular}
}
\caption{Comparison with baselines on the SoundSpaces-Speech for Seen and Unseen scenarios. Subjective (with 95\% confidence interval) and objective results with the different systems.}
\label{table:baseline}
\end{table*}

\subsection{Evaluation Metrics}
We measure the sample quality of the generated waveform using both objective metrics and subjective indicators. 
The objective metrics are designed to evaluate various aspects of waveform quality by comparing the ground-truth audio with the generated samples. 
Following the common practice of \citet{diffsinger, prodiff}, we randomly select 50 samples from the test set for objective evaluation. We provide three main metrics: (1) \textbf{Perceptual Quality}: This is assessed by human listeners using the Mean Opinion Score (MOS). A panel of listeners evaluates the audio's quality, naturalness, and its congruence with the accompanying image. Ratings are assigned on a scale from 1 (poor) to 5 (excellent). The final MOS is the average of these ratings. (2) \textbf{Room Acoustics (RT60 Error)}: RT60 measures the reverberation time in seconds for an audio signal to decay by 60 dB, which is a standard metric for characterizing room acoustics. To calculate the RT60 Error (RTE), we estimate the RT60 values from the magnitude spectrograms of the output audio, using a pre-trained RT60 estimator provided by \citet{vam}. (3) \textbf{Mel Cepstral Distortion (MCD)}: MCD quantifies the spectral distance between the synthesized and reference mel-spectrogram features. It is widely used as an objective measure of audio quality, particularly in tasks involving speech synthesis. Lower MCD values indicate higher spectral similarity between the generated and ground-truth audio.

Each of these metrics provides a distinct perspective on the quality of the generated waveform, allowing for a comprehensive evaluation of the system's performance.

\subsection{Baselines}

To demonstrate the effectiveness of our M$^2$SE-VTTS, we compare it against five baseline systems:

\begin{itemize}

\item \textbf{ProDiff} \cite{prodiff}: This first baseline is a progressive fast diffusion model designed for high-quality speech synthesis, where the input is text and the model directly predicts clean mel-spectrograms, significantly reducing the required sampling iterations.

\item \textbf{DiffSpeech} \cite{diffsinger}: This method is a TTS model that employs a diffusion probabilistic approach, where the input is text and the model iteratively converts noise into mel-spectrograms conditioned on the text.

\item \textbf{VoiceLDM} \cite{voiceldm}: The third system is a TTS model that uses text as its primary input, effectively capturing global environmental context from descriptive prompts to generate audio that aligns with both the content and the overarching situational description. Given the differences in environmental text descriptions between the training datasets—where the original dataset primarily describes the type of environment, while ours emphasizes the specific components and their spatial relationships—we choose to concentrate on the model's novel method of leveraging textual descriptions to guide the synthesis of reverberation speech during code reproduction.

\item \textbf{ViT-TTS-ResNet18} \cite{ViT-TTS}: The fourth baseline is a VTTS model that takes both text and environmental images as inputs, leveraging ResNet18 \cite{resnet} to extract global visual features from the image to enhance audio generation by capturing the room's acoustic characteristics.

\item \textbf{ViT-TTS-CLIP} \cite{ViT-TTS}: The last system is also ViT-TTS, which utilizes CLIP-ViT as a global RGB feature extractor.

\end{itemize}

\subsection{Main Results}

As shown in Table \ref{table:baseline}, the performance of the M$^{2}$SE-VTTS model on the test-unseen set is generally lower than that on the test-seen set, largely due to the presence of scenarios not encountered during training. 
Nevertheless, our model consistently outperforms all baseline systems across both sets, achieving the best results in RTE (0.0744), MCD (4.4215), and MOS (3.849 $\pm$ 0.025). 
These results demonstrate that our model is capable of synthesizing immersive reverberant speech. In addition, our model outperformed TTS diffusion models, such as DiffSpeech and ProDiff, across all metrics, notably in RTE. This indicates that traditional TTS models struggle to understand spatial environment information, focusing instead on audio content, pitch, and energy. To address this limitation, our multi-modal scheme learns more comprehensive spatial information. Furthermore, comparison with voiceLDM highlights the advantages of the multi-modal spatial cues and Gemini-based spatial environment understanding. Although voiceLDM takes environmental context descriptions as prompts to synthesize environmental audio, its choice of the spatial prompt and lack of a spatial semantic understanding strategy result in worse performance in predicting the correct reverberation and synthesizing high-quality audio with perceptual accuracy. Finally, ViT-TTS, which uses ResNet18 for global visual feature extraction, and ViT-TTS-CLIP, which employs CLIP-ViT, both outperform other baseline models. However, compared to our proposed model, both ViT-TTS and ViT-TTS-CLIP showed inferior performance in both test-unseen and test-seen environments. This suggests that our accurate modeling of the interaction between crucial local regions and the global context is effective, achieved by integrating knowledge gained from local spatial understanding.

In conclusion, our comprehensive evaluation results demonstrate the effectiveness of our proposed scheme in generating reverberant speech that matches the target environment.

\begin{table}[t!]
\centering

\setlength{\baselineskip}{2mm}{
\begin{tabular}{@{}l|ccc@{}}
\toprule
\textbf{System}       & \textbf{MOS} ($\uparrow$)               & \textbf{RTE}  ($\downarrow$)  & \textbf{MCD}  ($\downarrow$)  \\ \midrule
GT(voc.)              & 4.149 $\pm$ 0.027           & 0.0080          & 1.4600          \\ \midrule
w/o RGB               & 3.716 $\pm $ 0.049          & 0.0985          & 4.6378          \\
w/o Depth             & 3.753 $\pm $ 0.022          & 0.0957          & 4.6808          \\
w/o LLM               & 3.749 $\pm $ 0.026          & 0.0881          & 4.6121          \\
w/o LSU               & 3.753 $\pm $ 0.025          & 0.0984          & 4.6238          \\
w/o LGSU-L            & 3.698 $\pm $ 0.043          & 0.1011          & 4.6939          \\
w/o LGSU-G            & 3.703 $\pm $ 0.046          & 0.1039          & 4.7706          \\ \midrule
\textbf{M$^2$SE-VTTS} & \textbf{3.849 $\pm $ 0.025} & \textbf{0.0744} & \textbf{4.4215} \\ \bottomrule
\end{tabular}
}
\caption{\label{tab:wo}Ablation study results. 
The results of M$^2$SE-VTTS are sourced from Table \ref{table:baseline}.}
\end{table}

\begin{figure*}[ht!]
\centering
\centerline{
\includegraphics[width=1.035\linewidth]{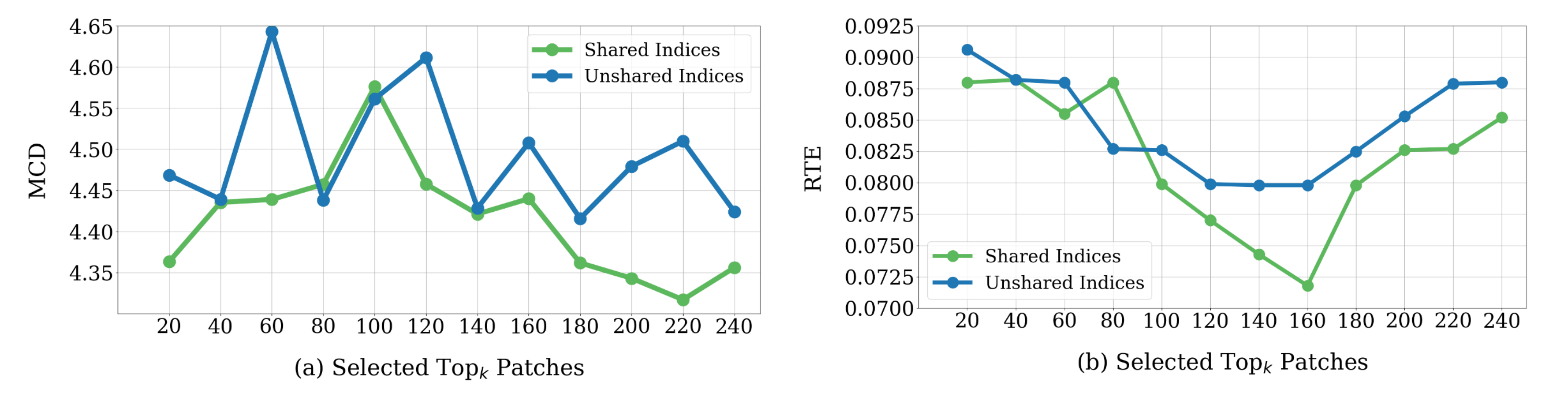}
}
\caption{The horizontal axis represents the Top$_k$ values, with the left plot showing MCD and the right plot displaying RTE. Green lines denote contribution parameter settings, while blue lines indicate comparison settings.}
\label{fig:topk}
\end{figure*}

\subsection{Ablation Results}

To evaluate the individual effects of several key techniques on the Test-Unseen set in our model, including the RGB space (RGB), the Depth space (Depth), the Gemini-based spatial semantic understanding (LLM), the local spatial understanding (LSU), local-aware interactions (LGSU-L), and global knowledge interactions (LGSU-G), we remove these components to build various systems. A series of ablation experiments were conducted, and the subjective and objective results are shown in Table \ref{tab:wo}.

We find that removing different types of modality information (w/o RGB and w/o Depth) in the Multi-modal Features Extraction led to a decrease in performance across most objective metrics, and the subjective MOS scores also dropped. This suggests that our multi-modal strategy can learn more comprehensive spatial information and enhance the expressiveness of reverberation.

In addition, to validate the Gemini-based spatial semantic understanding (LLM), we remove this component (w/o LLM). As shown in Table \ref{tab:wo}, the removal of the semantic understanding component led to a reduction in all subjective and objective metrics. This demonstrates that spatial images, when analyzed by Gemini for semantic understanding, enable our model to achieve a more accurate representation of reverberation.

Furthermore, we further explored the Top$_k$ region selection for RGB/Depth (w/o LSU). Omitting these critical regions leads to a decrease in both subjective and objective metrics. This suggests that by identifying important semantic information, the model can accurately understand the spatial environment and improve the style and quality of the reverberation.

Finally, we remove local semantics (w/o LGSU-L) and global context (w/o LGSU-G) in the Local-aware Global Spatial Understanding component. This removal results in decreased performance across both subjective and objective metrics, underscoring the efficacy of our multi-modal and multi-scale approach in modeling the interplay between local and global spatial contexts for reverberation.

\subsection{Top$_k$ Index Sharing Comparative Study}

To evaluate the effectiveness of selecting depth features using shared Top$_k$ indices from the RGB image and to compare this approach with independent semantic-guided methods, we focus on the efficacy of index-sharing and the impact of different Top$_k$ values. Specifically, we design experiments to compare two feature selection strategies: shared Top$_k$ indices and unshared Top$_k$ indices.

In the shared indices strategy, Top$_k$ critical regions from RGB images guide depth feature selection. In contrast, the unshared Top$_k$ indices strategy independently selects features from RGB and depth images based on their respective semantic information. Various Top$_k$ values (e.g., 20, 40, ..., 240) are tested to observe their effects on performance, evaluated using two objective metrics, with lower values indicating better performance.

The experimental results demonstrate that the shared Top$_k$ strategy consistently outperforms the unshared Top$_k$ strategy for all tested values of Top$_k$. This approach yields lower values for the objective metrics, indicating more natural audio generation that better aligns with the environment. As the Top$_k$ value increases, performance improves as more comprehensive spatial information is captured, though it plateaus or declines slightly after reaching a threshold (e.g., 140).

The comparison of strategies confirms that shared Top$_k$ indices from RGB more effectively capture critical spatial information, leading to more realistic and environment-consistent audio. This suggests that the shared index strategy is superior for depth feature selection in multi-modal tasks, providing guidance for improving feature selection in future multi-modal systems. Further refinement of this index-sharing approach is recommended to maximize performance.

\section{Conclusion}

This paper introduces M$^2$SE-VTTS, an innovative multi-modal and multi-scale approach for Visual Text-to-Speech (VTTS) synthesis. Our method addresses the limitations of previous VTTS systems, such as their limited spatial understanding, by incorporating both RGB and depth images to achieve a comprehensive representation of the spatial environment. 
This comprehensive spatial representation includes modeling both local and global spatial contexts, which are crucial for capturing the environmental nuances influencing speech reverberation. By combining local spatial understanding guided by the environment caption and local-aware global spatial modeling, M$^2$SE-VTTS effectively captures the interactions between different spatial scales, which are essential for accurate reverberation modeling. Evaluations demonstrate that our model consistently outperforms state-of-the-art benchmarks in generating environmental speech, establishing a new standard for environmental speech synthesis in VTTS.

Despite its advances, the M$^2$SE-VTTS framework has limitations, such as increased computational complexity from multi-modal and multi-scale feature integration, potentially hindering real-time applications. Additionally, the model's performance is inconsistent in unseen environments, highlighting the need for improved generalization. Future research should focus on optimizing computational efficiency and enhancing the model's adaptability to unseen spatial contexts.

\section{Acknowledgments}

This work by Rui Liu was funded by the Young Scientists Fund (No. 62206136) and the General Program
(No. 62476146) of the National Natural Science Foundation
of China, 
and the ``Inner Mongolia Science and Technology Achievement Transfer and Transformation Demonstration Zone, University Collaborative Innovation Base, and University Entrepreneurship Training Base'' Construction Project (Supercomputing Power Project) (No.21300-231510). The research by Haizhou Li was partly supported by Shenzhen Science and Technology Program (Shenzhen Key Laboratory Grant No. ZDSYS20230626091302006) and  Shenzhen Science and Technology Research Fund (Fundamental Research Key Project Grant No. JCYJ20220818103001002).

\bibliography{aaai25}

\end{document}